%% file: final.tex
\documentclass{article}

\usepackage[final]{neurips_2019}

\usepackage[utf8]{inputenc} 
\usepackage[T1]{fontenc}    
\usepackage{hyperref}       
\usepackage{url} 
\usepackage{subfig}
\usepackage{float}
\usepackage{booktabs}       
\usepackage{amsfonts}       
\usepackage{nicefrac}       
\usepackage{microtype}      
\usepackage[textsize=tiny]{todonotes}
\input{macro}

\title{Reproducibility Report: Test-Time Training on Nearest Neighbors for Large Language Models}

\author{%
  Boyang Zhou, Johan Lindqvist, Lindsey Li \\ 
 \texttt{\{zby2003, lindq2, linjli\}@cs.washington.edu} \\ 
}

\begin{document}

\maketitle



\section*{\centering Reproducibility Summary}


\subsection*{Scope of Reproducibility}

The central claim we tried to reproduce from \citep{hardt2024test} is that fine-tuning a pretrained language model on 20 retrieved nearest neighbors per test input—using a single gradient update per neighbor—significantly reduces perplexity and narrows the performance gap between smaller and larger models. We verify this claim on several Pile tasks using GPT-2~\citep{gpt2} (117M and 774M) and GPT-Neo~\citep{gpt-Neo} (1.3B).

\subsection*{Methodology}


We reproduced the test-time training pipeline by reusing pretrained RoBERTa~\citep{roberta} embeddings finetuned on The Pile dataset~\citep{gao2020pile800gbdatasetdiverse}, indexed with Faiss~\citep{faiss} using a Flat L2 index. For each test sequence, we retrieved 20 nearest neighbors and fine-tuned four models (GPT-2 117M, GPT-2 774M, GPT-Neo 1.3B, and R1-Distilled-Qwen2.5-1.5B~\citep{deepseekai2025deepseekr1incentivizingreasoningcapability}) with a single gradient update per neighbor. We used The Pile dataset from Hyak and adapted the authors' GitHub code (\url{https://github.com/socialfoundations/tttlm}) for our reproduction. Experiments ran on various GPUs, with evaluation times ranging from 1 hour to 38 hours. We evaluated using bits per byte, byte perplexity, and word perplexity metrics~\citep{lm-evaluation-harness}.

\subsection*{Results}


Our experiments successfully reproduced the key claims from the original paper. TTT-NN significantly reduces perplexity across diverse tasks, with strongest improvements on specialized domains like GitHub (51\% of original bits per byte) and EuroParl (68\%). Models not pre-trained on The Pile benefit more from TTT-NN than those already trained on this data. We extended evaluation to R1-Distilled-Qwen-1.5B, confirming TTT-NN's effectiveness for modern reasoning models. While the exact metric numbers differ slightly from the original paper due to using fewer nearest neighbors (20 vs 50), the overall patterns and conclusions remain consistent.

\subsection*{What was Easy}


The pretrained RoBERTa embeddings finetuned on The Pile dataset were readily accessible for download, and starting the indexing server with one split of the Pile dataset was simple when using the same system configuration. Despite The Pile dataset being removed from Hugging Face and its official website, we were able to locate a copy on Hyak, and the provided evaluation scripts made it easy to replicate the evaluation pipeline.

\subsection*{What was Difficult}

Reproducing the full-scale experiment proved challenging due to resource constraints. The paper implemented 30 CPU servers for retrieval from the entire Pile dataset, while the codebase required over 128GB RAM per server—exceeding our available allocation. We optimized the code to load only line locations rather than entire files, reducing memory usage to 32GB per server but increasing retrieval times. Training time remained our primary bottleneck, limiting our evaluation to select subsets of the data.



\subsection*{Communication with Original Authors}

We did not contact the authors as the paper and code were sufficiently clear and well-documented to enable our reproduction efforts without requiring direct communication.

\section{Introduction}

This project aims to reproduce the method proposed in  \textit{Test-Time Training on Nearest Neighbors for Large Language Models}  \citep{hardt2024test}. The paper introduces Test-time Training with Nearest Neighbors (TTT-NN), a simple yet powerful technique that fine-tunes a pretrained language model at test time using retrieved nearest-neighbor examples from a large-scale database (the Pile dataset~\citep{gao2020pile800gbdatasetdiverse}). Instead of incorporating retrieved data directly into the model’s input context, the approach performs a single gradient update per neighbor, typically using around 20 neighbors per test instance. This quick, on-the-fly adaptation is designed to locally refine the model’s predictions, leading to significant reductions in perplexity across a variety of language modeling tasks.

The authors hypothesize that test-time training on these retrieved neighbors not only improves performance on unseen tasks (such as code generation in \textit{pile\textunderscore github}) but also narrows the performance gap between smaller and larger models. In their experiments, they evaluate this hypothesis on 22 tasks from the Pile dataset using models of varying sizes: GPT-2 (117M parameters), GPT-2 Large (774M parameters)~\citep{gpt2}, and GPT-Neo~\citep{gpt-Neo} (1.3B parameters). The results indicate that TTT-NN can yield perplexity reductions of up to about 70\% on code tasks when using between 20 and 50 neighbors, with a smaller GPT-2 model enhanced by TTT-NN approaching the performance of a model over ten times larger.


\section{Scope of Reproducibility}




The original paper presents several significant claims:

\begin{enumerate}
\item Metrics such as \textit{bits per byte}, \textit{word perplexity}, and \textit{byte perplexity} significantly decrease after just a single gradient iteration on as few as 50 nearest neighbors. Particularly notable is that more than half of this performance gain occurs after training with just the 20 closest neighbors, an effect especially pronounced in tasks such as the \textit{pile\_github} subset of The Pile dataset.

\item The authors assert that models not pre-trained on a specific task can achieve comparable performance to models explicitly pre-trained, after training on up to 50 nearest neighbors during test time, when evaluated using the \textit{bits per byte} metric.

\item Training sequentially, from the nearest neighbor to the farthest, yields better results than training in reverse order (from farthest to nearest). This suggests that proximity-based ordering plays a significant role in model performance.

\item Training on 50 nearest neighbors at test time reportedly outperforms alternative methods, including: (a) in-context learning with neighbors, (b) interpolation of the token distribution among neighbors \citep{Khandelwal2020Generalization}, and (c) dynamic evaluation with multiple iterations \citep{pmlr-v80-krause18a}.
\end{enumerate}

In this project, we specifically address the first two claims by reproducing and verifying the experiments described in the original study. The remaining two claims were not directly supported by the available codebase provided by the authors or would have required training additional new models. Consequently, our analysis is focused primarily on validating the initial two claims in depth.

\subsection{Model Descriptions}

The original paper \citep{hardt2024test} introduces a novel method to enhance language model performance through test-time fine-tuning. Instead of relying solely on a static pre-trained model during inference, this approach dynamically adapts the model for each input by retrieving similar sequences from a pre-computed nearest neighbor structure and fine-tuning on these examples before generating a prediction. This context-aware adaptation mechanism significantly improves performance across various language modeling tasks.

The authors employed a carefully chosen suite of language models. RoBERTa \citep{roberta}, a bidirectional transformer architecture, was utilized to generate high-quality sequence embeddings, facilitating effective similarity search. For the test-time training experiments, the authors selected three different auto-regressive generative models with increasing model capacities: two GPT-2 variants \citep{gpt2} and the larger GPT-Neo model \citep{gpt-Neo}, enabling evaluation of how the proposed method scales with model size. All variants of GPT-2 were not pretrained on The Pile dataset, while GPT-Neo was pretrained on the entire The Pile dataset.

To construct the retrieval system, the authors first fine-tuned a pre-trained RoBERTa model on The Pile dataset—a diverse 800GB collection of text from 22 sources—over approximately 1.7 million training iterations. This specialized model then embedded all sequences from The Pile~\citep{gao2020pile800gbdatasetdiverse} into a high-dimensional vector space, where semantic similarity corresponds to geometric proximity. For efficient nearest neighbor search within this embedding space, the authors used the Faiss library \citep{faiss}, specifically employing a Flat L2 Index that performs exact distance calculations to ensure optimal retrieval quality.

The test-time training procedure was systematically evaluated using GPT-2 variants with 117M and 774M parameters, and GPT-Neo with 1.3B parameters. During inference, each validation sequence is embedded using the fine-tuned RoBERTa model, which queries the Faiss index to retrieve the $k$-nearest neighbors from The Pile. The language model is subsequently fine-tuned on these retrieved examples through one gradient update before evaluation.

A key advantage of this approach is that the fine-tuning process utilizes identical training objectives (i.e., language modeling loss) and hyperparameters as those originally used during model training, requiring no additional parameter tuning or specialized adaptation mechanisms. Thus, the method is readily applicable to existing language models without extensive modifications. The authors evaluated model performance using standard language modeling metrics, such as bits per byte (a measure of compression efficiency), byte perplexity, and word perplexity \citep{lm-evaluation-harness}, demonstrating consistent improvements across all model sizes and evaluation criteria.

As an additional study we also evaluate the TTT-NN framework for a recent state-of-the-art reasoning model of comparable size, the R1-Distilled-Qwen2.5-1.5B ~\citep{deepseekai2025deepseekr1incentivizingreasoningcapability} model. This model was chosen to study the effects of the nearest neighbor training on a more modern architecture and better performance before test time training. Qwen2.5 1.5B also differs from the other models in that it is a reasoning model, skewing it to perform well on longer reasoning tasks, meaning that the objective is to perform well on tasks like mathematical problem-solving, logical deduction, and extended analysis. This is out-of-distribution of the Pile dataset in this study which focused on general text completion. Our evaluation aims to gain an understanding whether TTT-NN may still improve performance in this context. 
We specifically chose to evaluate this model on a math-focused task, as the model was trained especially to achieve good performance on math understanding potentially implying that any improvements in performance may be more difficult to achieve. 
Although being an open source model, R1-Distilled-Qwen2.5-1.5B did not disclose its training data so we do not know whether it was pretrained on The Pile dataset.


\subsection{Datasets}

The dataset used for the reproducing this study is The Pile \citep{gao2020pile800gbdatasetdiverse}, an extensive and open-source corpus curated by EleutherAI, specifically designed to support the training of large language models (LLMs). The Pile aggregates text from diverse domains into 22 carefully selected sub-datasets, collectively comprising 825 GiB of English textual data in its raw form. To enhance the representation of specific subsets deemed highly valuable or relevant, a version of The Pile employs upsampling techniques, expanding the dataset to an effective size of approximately 1,254 GiB. Access to The Pile for this project was obtained through Hyak, which hosts a complete copy of the dataset \citep{hyakPile}.

\begin{table}[tbh]
    \centering
    
    \small
    \tablestyle{8pt}{1.1}
    \begin{tabular}{|l|l|l|r|r|r|r|}
    \hline
    Dataset Name & Size & Description & GPT2 & GPT2-large & GPT-Neo & \makecell[l]{R1-Distill-\\Qwen-1.5B} \\
    & & & 117M & 774M & 1.3B & 1.5B\\
    \hline 
    Wikipedia & 6.85 GB & \makecell[l]{Encyclopedia articles\\from English Wikipedia} & 
\checkmark & - & - & -	\\
\hline
    Arxiv & 60.36 GB & \makecell[l]{Academic papers\\covering physics, mathematics,\\ computer science,\\and other scientific disciplines} & 
\checkmark & - & - & - \\
\hline
    Books3 & 108.40 GB & \makecell[l]{A large collection of\\books from various sources} & 
\checkmark & - & - & - \\
\hline
    Enron emails & 0.95 GB & \makecell[l]{Corporate email messages\\ from the Enron Corporation} & 
\checkmark & - & - & - \\
\hline
    DM Math & 8.32 GB & \makecell[l]{DeepMind Mathematics\\ dataset containing\\ mathematical content} & 
\checkmark & 
\checkmark & 
\checkmark & 
\checkmark\\
\hline
    EuroParl & 4.93 GB & \makecell[l]{Proceedings of the European\\ Parliament in multiple languages} & 
\checkmark & 
\checkmark & 
\checkmark & 
\makecell[l]{Crash due to\\ OOM}\\
\hline
    Github & 102.18 GB & \makecell[l]{Source code from \\public GitHub repositories} & 
\checkmark & 
\checkmark & 
\makecell[l]{Partial \\(~80\%)} & - \\
\hline
    
    \end{tabular}
    \caption{Details on sub-datasets of The Pile that were used in our experiments across different models.}
    \label{tab:datasets}
\end{table}

Due to limitations in compute and time constraints we were unable to evaluate TTT-NN on all of the sub-datasets for all models. Particularly for some of the larger models and datasets evaluation time was unfeasibly long. Table~\ref{tab:datasets} shows the details about the datasets used and for which models the datasets were evaluated for.

For the evaluations on GPT2 the datasets were chosen with regards to diversity in terms of the type of data, average training cost and size of the datasets. As GPT2 was the smallest model, it was able to run faster than the others which meant that it was possible to evaluate more of the larger datasets. The larger models posed more limitations in terms of what datasets could be evaluated and as such we focused our efforts on a subset of datasets which the authors claimed saw the greatest gains in performance while also taking into account the diversity in terms of content.


\subsection{Hyperparameters}

We carefully selected hyperparameters based on the original paper and its publicly available codebase. The hyperparameters used in our experiments are summarized below:

\begin{itemize}[leftmargin=*]
\item \textbf{Batch Size:} Set to 16, as specified in the authors' publicly released code.
\item \textbf{Number of Gradient Iterations:} One iteration, consistent with the paper.
\item \textbf{Number of Nearest Neighbors:} We employed 20 nearest neighbors for test-time training, the reduced number as compared to the 50 used in the original paper stems from concerns related to runtime limits.
\item \textbf{Training Order:} Neighbors were trained sequentially in ascending order of distance, starting with the nearest neighbor.
\item \textbf{Max length and Stride:} Set to 1024 for GPT2 and 2048 for GPT2-Neo, GPT2-Large and R1-distilled Qwen1.5B, the same default parameters as used by the authors. 
\item \textbf{Learning Rate:} Set to 2e-5 for GPT2 and 5e-6 for GPT2-Neo, GPT2-Large and R1-distilled Qwen1.5B, the same default parameters as used by the authors. 
\end{itemize}


\subsection{Implementation}


We utilized the official codebase provided by the authors at \hyperlink{https://github.com/socialfoundations/tttlm}{https://github.com/socialfoundations/tttlm}. This repository includes scripts for training the embedding model, building the nearest neighbor index on The Pile dataset, running distributed servers for the index, querying these servers, evaluating TTT-NN on The Pile, and executing baseline comparisons. The code is primarily written in Python and leverages several packages, including PyTorch for model training, FAISS for efficient similarity search, and Hugging Face's Transformers library for handling pre-trained models. Detailed instructions for setting up the environment, including dependencies and configurations, are provided in the repository's README file.

To address resource constraints, particularly RAM usage during indexing, we introduced modifications to the indexing pipeline. Specifically, rather than loading entire datasets into memory, our implementation loads only the offset in byte into the memory, and store a "pointer" to the location in storage. This optimization significantly reduces memory requirements and enhances efficiency, particularly for systems with limited RAM. We recommend authors and practitioners consider similar data-loading strategies to improve ram efficiency when dealing with loading relationships between large sets of tokens. We have made our modified code publicly available at \hyperlink{https://github.com/boezzz/tttlm/tree/file_indexing}{https://github.com/boezzz/tttlm}, with commit histories that clearly outlines the changes.  

\subsection{Experimental Setup}

Our experiments were run using the test-time training pipeline from the authors' GitHub repository, with modifications to address memory efficiency concerns.
We executed our experiments across a variety of GPU resources based on availability, including NVIDIA 2080Ti, A40, L40, and A100 GPUs.
For each experiment, we followed these steps:

\begin{itemize}[leftmargin=*]
    \item Started an indexing server with a single split of The Pile dataset to provide nearest neighbor retrieval capabilities
\item For each test sequence in the evaluation dataset:
\begin{itemize}
    \item Embedded the sequence using the fine-tuned RoBERTa model
Retrieved 20 nearest neighbors from The Pile using the Faiss index
\item Fine-tuned the language model (GPT-2, GPT-2 Large, GPT-Neo, or R1-Distilled-Qwen) with a single gradient update per neighbor
\item Evaluated the performance using bits per byte, byte perplexity, and word perplexity metrics; 
\end{itemize}
\end{itemize}
Our modified code and experiment notebooks are publicly available at \hyperlink{https://github.com/boezzz/tttlm/tree/file_indexing}{https://github.com/boezzz/tttlm}, with detailed commit histories documenting our changes to the original implementation.

\subsection{Computational Requirements} 

Test-time fine-tuning for each test sequence represents the primary GPU computational cost of our experiments. To maintain practical inference times, we evaluated a subset of The Pile’s test set, mirroring the approach described by the original authors. In the original paper, each evaluation required approximately 1.35 seconds per sequence for retrieval, and then the retrieved sample is subsequently used for gradient update steps before evaluation. Initially, we estimated reproducing key results for GPT-2 models on selected tasks would take around 2–3 days on a single A100 GPU, including setup and iterative debugging. In practice, our fine-tuning experiments were executed using different GPUs (the first available ones that can fit the model finetuning), as listed in Table~\ref{tab:training_times}. The training times generally exceeded our expectations largely due to the usage of less powerful GPUs and a general underestimation of the evaluation task of some of the larger datasets. The evaluation of pile github on GPT-Neo  was not fully completed due to resource allocation time limits, which meant the run had to be canceled early. The partial results are used in the results section.

For indexing and retrieval, the original authors utilized 30 server instances, each configured with 12 CPUs and 256 GB of RAM, totaling 360 CPUs and approximately 7.68 TB of RAM. Each server served six replicas leading to a total of 180 instances. In contrast, our optimized implementation can host these indexes on 30 servers with each servers using 2 CPUs and 32 GB of RAM though only running a single instance, totaling 60 CPUs and 960 GB of RAM. Due to improvements—specifically, selectively loading only necessary lines from datasets instead of entire files—each indexing engine required significantly less memory than the original 128 GB estimate per server. This optimization substantially reduced overall computational resource demands at the cost of slower retrieval. The querying in the original paper setup had a mean query time of 1.35s while we observe a 5s average query time. However, the querying process is only a very insignificant portion of the total training time per iteration for most of the evaluations, as such optimizing the efficiency of retrieving the nearest neighbors would only give us marginal gain of few seconds.

\begin{table}[tbh]
    \centering
    \small
    \tablestyle{4pt}{1.1}
    \begin{tabular}{|l|l|l|r|r|r|}
        \hline
        Dataset & Model & GPU & Iterations & Total Run Time & Seconds/Iteration \\
        \hline
        Wiki & GPT2 (117M) & 2080Ti & 3502 & 12:34:57 & 12.93 \\
        \hline
        Arxiv &  GPT2 (117M)  & 2080Ti & 481 & 14:10:38 & 106.11 \\
        \hline
        Books3 &  GPT2 (117M)  & 2080Ti & 53 & 10:47:52 & 733.44 \\
        \hline
        Enron &  GPT2 (117M)  & 2080Ti & 202 & 1:13:58 & 21.97 \\
        \hline
        DMMath &  GPT2 (117M)  & 2080Ti & 384 & 2:44:50 & 25.76 \\
        DMMath & GPT2-Large (774M) & A40 & 384 & 7:54:27 & 73.17 \\
        DMMath & GPT-Neo (1.3B) & A40 & 384 & 9:53:28 & 92.73 \\
        DMMath & R1-Distill-Qwen-1.5B & A100 & 384 & 13:49:18 & 129.58 \\
        \hline
        Euro & GPT2 (117M) & 2080Ti & 31 & 1:31:50 & 177.74 \\
        Euro & GPT2-Large (774M) & A40 & 31 & 4:06:37 & 477.35 \\
        Euro & GPT-Neo  (1.3B) & A40 & 31 & 3:20:25 & 387.91 \\
        \hline
        Github & GPT2 (117M) & 2080Ti & 3639 & 21:40:43 & 21.45 \\
        Github & GPT2-Large (774M) & L40 & 3639 & 33:22:33 & 33.02 \\
        Github & GPT-Neo (1.3B) & L40 & 2867/3639 & 38:12:33 (Projected Remaining, 21:25:01) & 99.87 \\
        \hline
    \end{tabular}
    \caption{Computational resources and training times for evaluations}
    \label{tab:training_times}
\end{table}



\section{Results}
In this section, we present the results of our reproducibility study for Test-Time Training with Nearest Neighbors (TTT-NN). Our experiments provide insights into both the computational requirements and performance improvements of this approach.

\begin{figure}[tbh]
    \centering
    \includegraphics[width=0.95\textwidth]{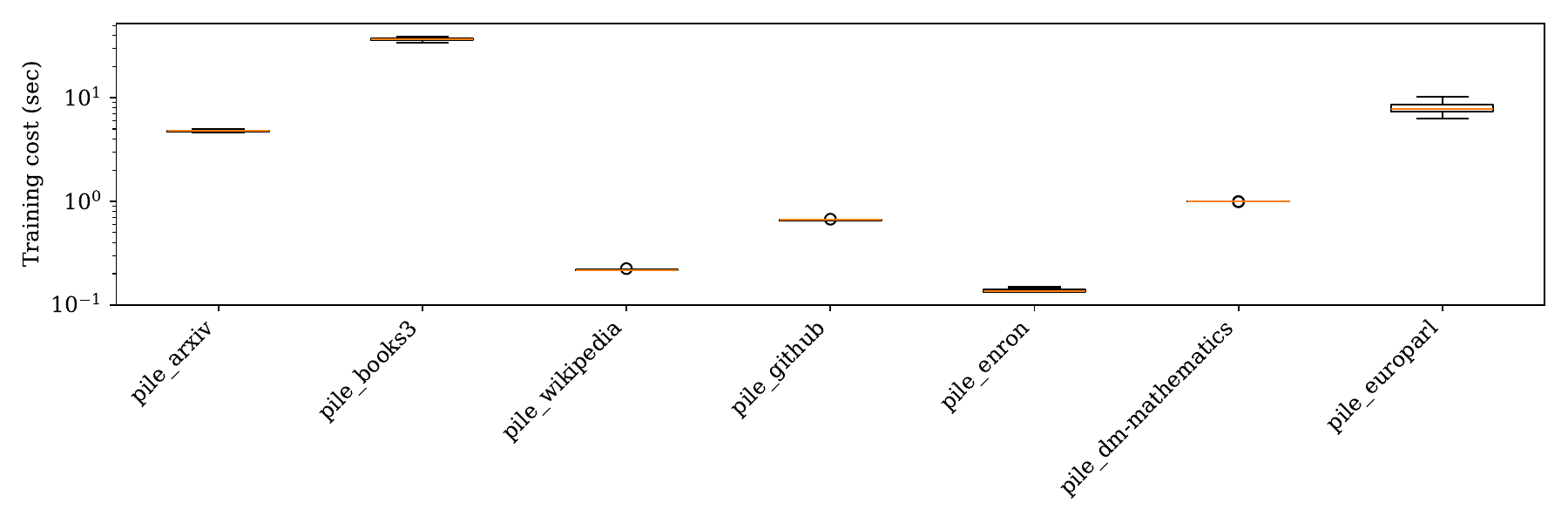}
    \caption{Training cost in seconds per neighbor using GPT2}
    \label{fig:training_time_per_neighbor}
\end{figure}

The computational demands of TTT-NN represent a significant consideration for practical deployment. As shown in Table~\ref{tab:training_times}, the total training time and the time needed for each training iteration (per testing sample) vary substantially across tasks and models, given our constrained computational resources. For larger models like GPT-Neo and R1-Distilled-Qwen2.5-1.5B, each test instance requires approximately 100 seconds on DM-Math dataset under our resource limitations, creating a substantial computational overhead for real-time applications. It's important to note that the seconds per iteration reported in Table~\ref{tab:training_times} depends on the number of neighbors we retrieve, which we used  20 for all our experiments.  In Figure~\ref{fig:training_time_per_neighbor}, we observed substantial variation in training time per nearest neighbor across different datasets with the same GPT-2 model and the same 2080Ti GPU. Some datasets, particularly Books3, required significantly more computational time per iteration (over 36 seconds) compared to others like Euro (less than 1 second).  This variability demonstrates that training costs for TTT-NN are highly dataset-dependent, which has important implications for practical deployment across different domains. It is worth noting that with more powerful infrastructure, as used in the original paper, these times could potentially be reduced. Nevertheless, the scaling relationship we observed has important implications for the practical applicability of TTT-NN in production environments where latency is a concern and computational resources may be limited.

We now evaluate the two primary claims from the original paper: (1) that metrics such as bits per byte and perplexity significantly decrease after fine-tuning on nearest neighbors (Sec.~\ref{sec:res1}), and (2) that models not pre-trained on specific tasks can achieve comparable performance to pre-trained models after test-time training (Sec.~\ref{sec:res2}). Finally, we present additional results extending beyond the original paper, including experiments with the more recent R1-Distilled-Qwen2.5-1.5B model to assess how well the method generalizes to contemporary architecture designs (Sec.~\ref{sec:add_res}).

\begin{figure}[tbh]
    \centering
    \subfloat{
        \includegraphics[width=0.13\textwidth]{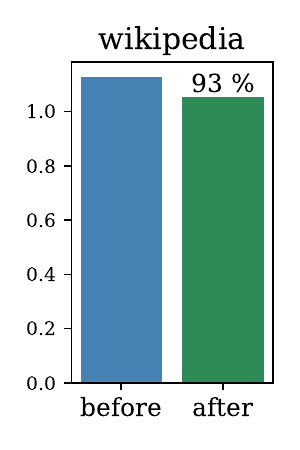}
        \label{fig:gpt-2wiki-perplexity}
    }
    \subfloat{
        \includegraphics[width=0.13\textwidth]{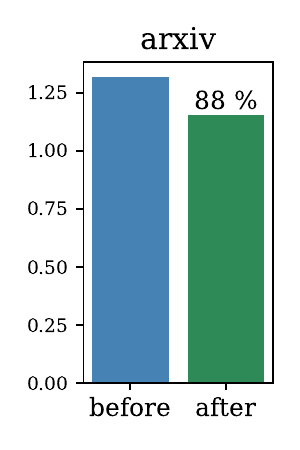}
        \label{fig:gpt2-arxiv-perplexity}
    }
    \subfloat{
        \includegraphics[width=0.13\textwidth]{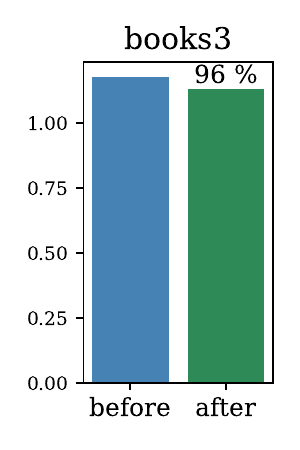}
        \label{fig:gpt2-book3-perplexity}
    }
    \subfloat{
        \includegraphics[width=0.13\textwidth]{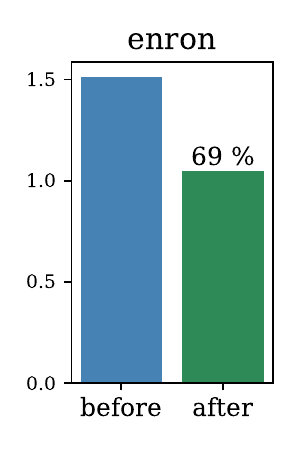}
        \label{fig:gpt2-enron-perplexity}
    }
    \subfloat{
        \includegraphics[width=0.13\textwidth]{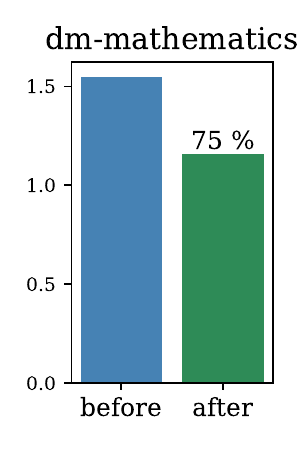}
        \label{fig:gpt2-dm-perplexity}
    }
    \subfloat{
        \includegraphics[width=0.13\textwidth]{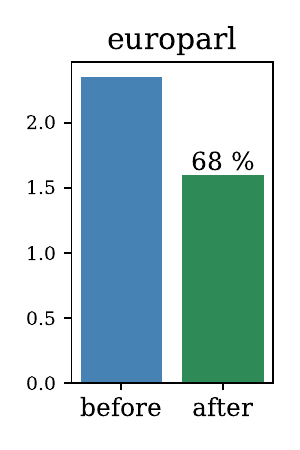}
        \label{fig:gpt2-euro-perplexity}
    }
    \subfloat{
        \includegraphics[width=0.13\textwidth]{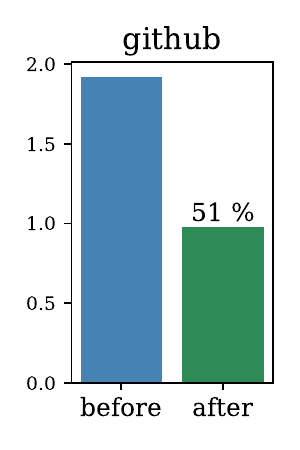}
        \label{fig:gpt2-github-perplexity}
    }

    \caption{Result comparison between before and after applying TTT-NN with 20 nearest neighbors on GPT-2 on 7 sub-datasets from the Pile, including wikipedia, arxiv, books3, enron, dm-mathematics, europarl and github. The metric used here is bits per byte.}
    \label{fig:gpt2-before-after}
\end{figure}

\begin{figure}[tbh]
    \centering
    \subfloat{
        \includegraphics[width=0.13\textwidth]{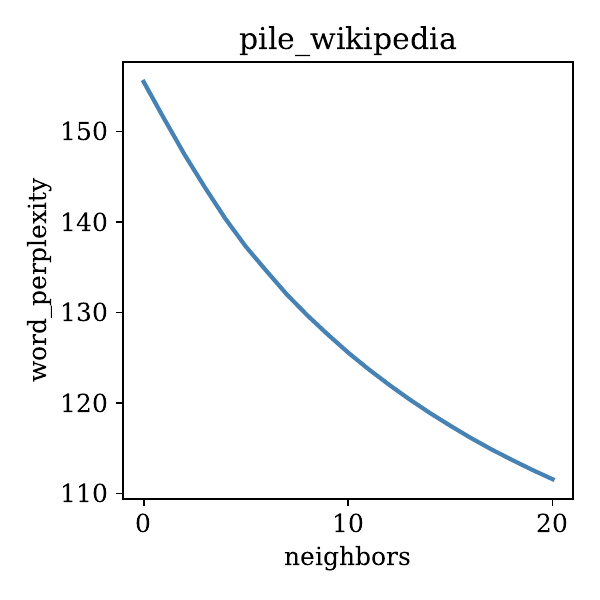}
        \label{fig:gpt-2wiki-perplexity}
    }
    \subfloat{
        \includegraphics[width=0.13\textwidth]{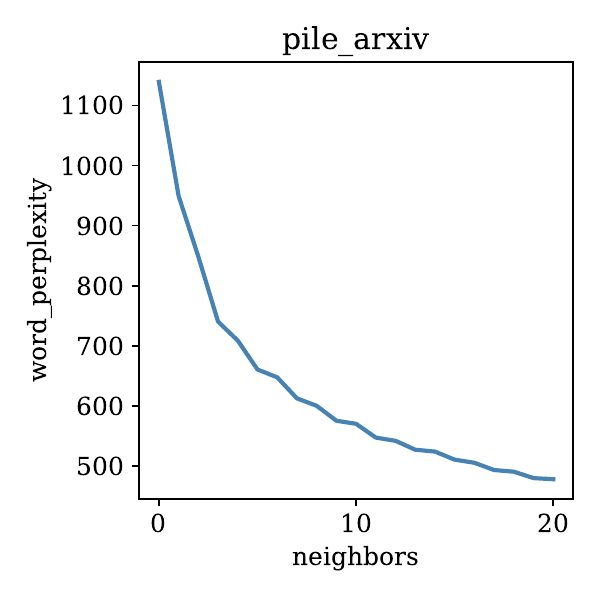}
        \label{fig:gpt2-arxiv-perplexity}
    }
    \subfloat{
        \includegraphics[width=0.13\textwidth]{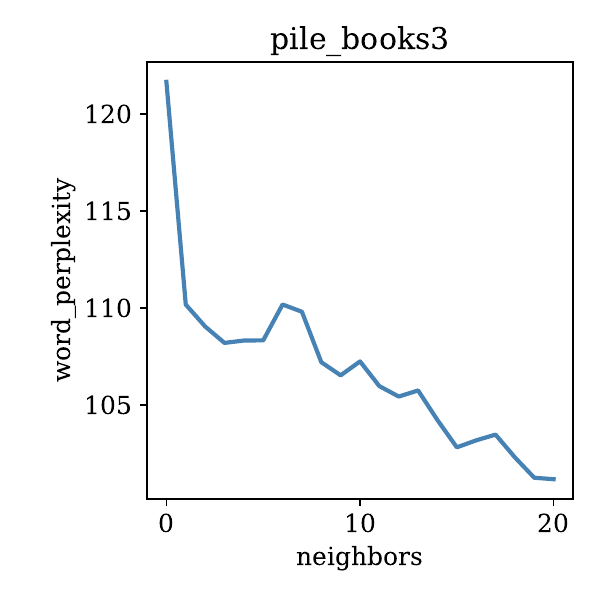}
        \label{fig:gpt2-book3-perplexity}
    }
    \subfloat{
        \includegraphics[width=0.13\textwidth]{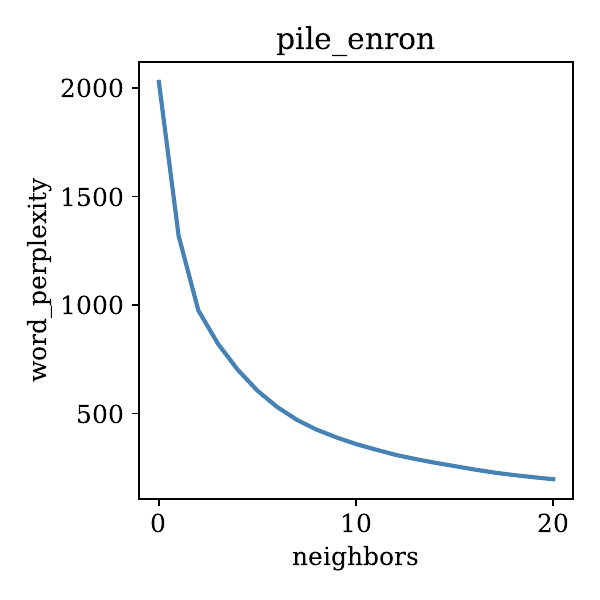}
        \label{fig:gpt2-enron-perplexity}
    }
    \subfloat{
        \includegraphics[width=0.13\textwidth]{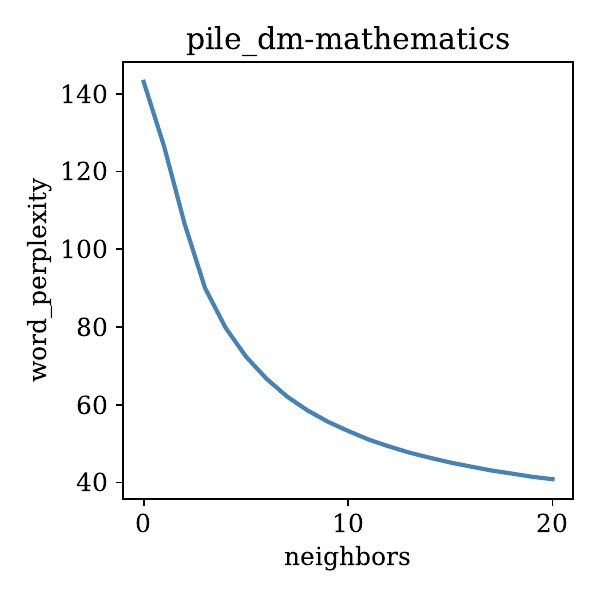}
        \label{fig:gpt2-dm-perplexity}
    }
    \subfloat{
        \includegraphics[width=0.13\textwidth]{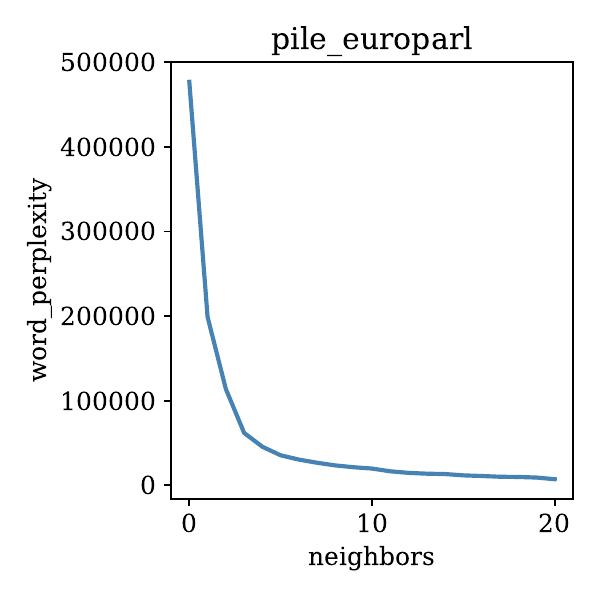}
        \label{fig:gpt2-euro-perplexity}
    }
    \subfloat{
        \includegraphics[width=0.13\textwidth]{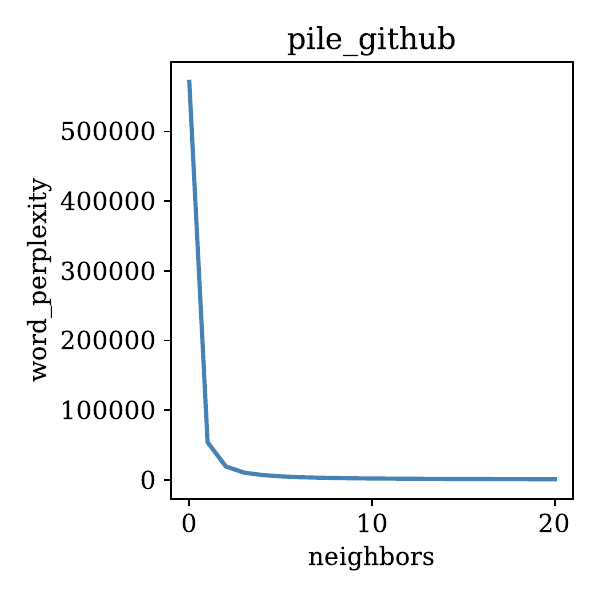}
        \label{fig:gpt2-github-perplexity}
    }

    \subfloat{
        \includegraphics[width=0.13\textwidth]{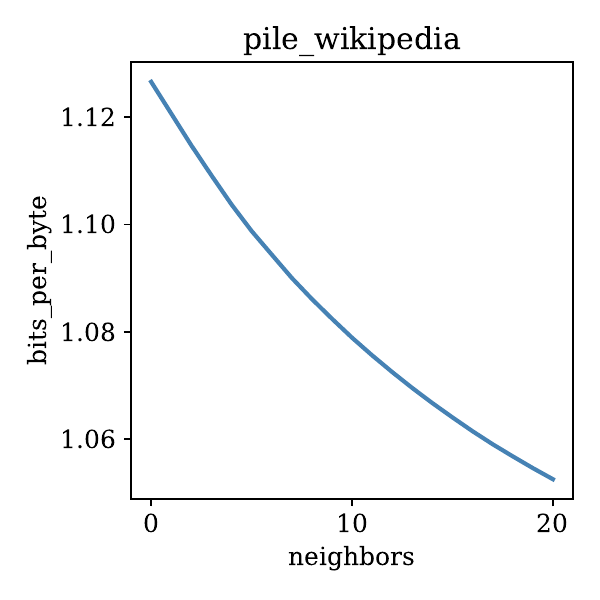}
        \label{fig:gpt-2wiki-bpb}
    }
    \subfloat{
        \includegraphics[width=0.13\textwidth]{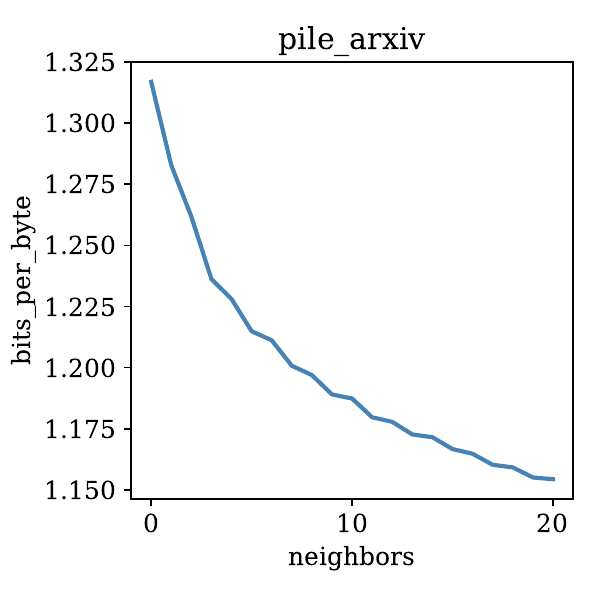}
        \label{fig:gpt2-arxiv-bpb}
    }
    \subfloat{
        \includegraphics[width=0.13\textwidth]{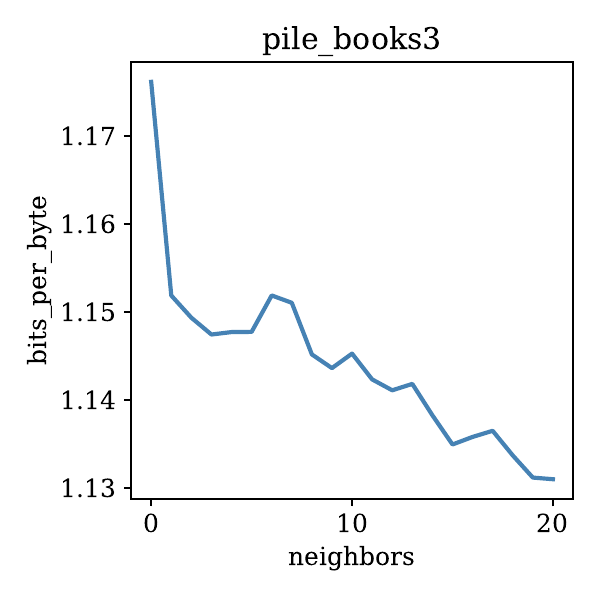}
        \label{fig:gpt2-book3-bpb}
    }
    \subfloat{
        \includegraphics[width=0.13\textwidth]{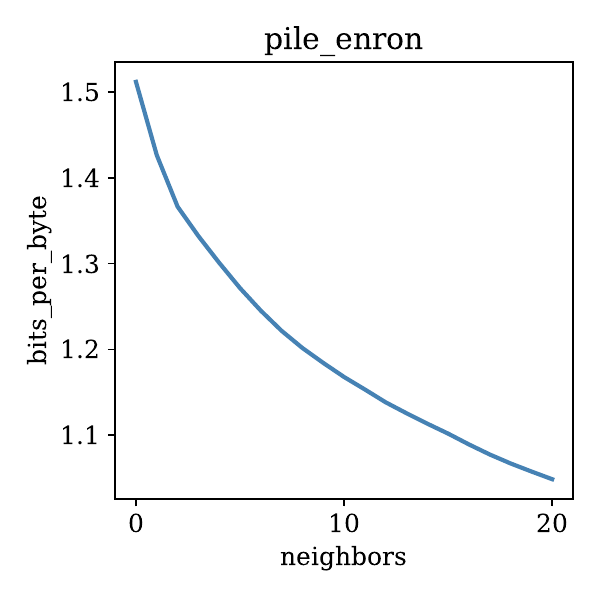}
        \label{fig:gpt2-enron-bpb}
    }
    \subfloat{
        \includegraphics[width=0.13\textwidth]{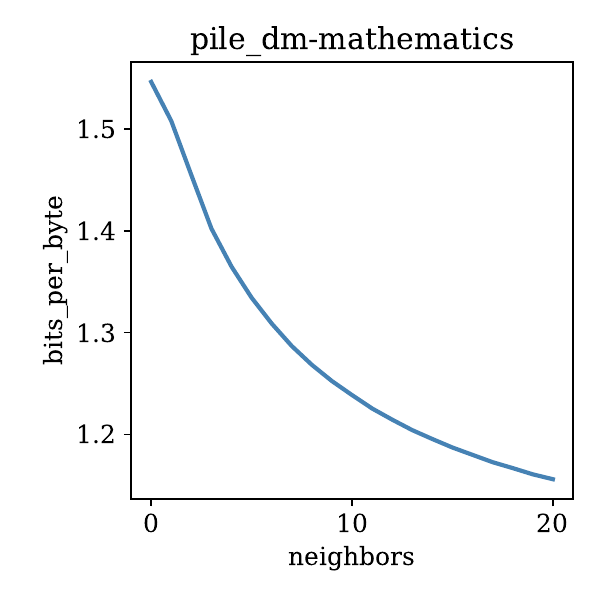}
        \label{fig:gpt2-dm-bpb}
    }
    \subfloat{
        \includegraphics[width=0.13\textwidth]{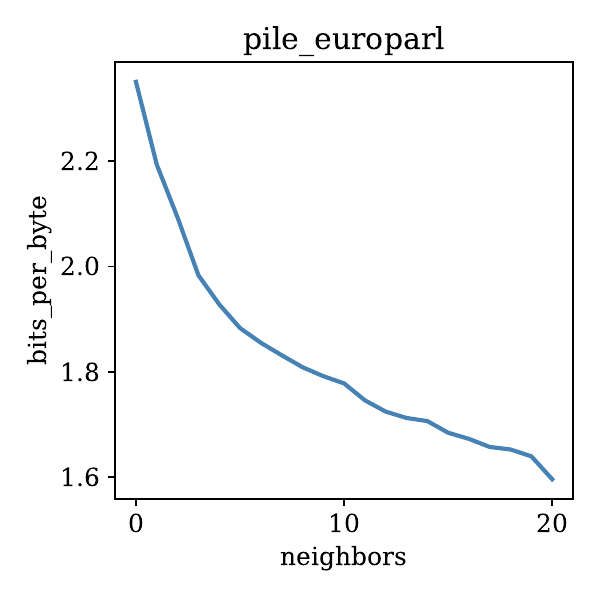}
        \label{fig:gpt2-euro-bpb}
    }
    \subfloat{
        \includegraphics[width=0.13\textwidth]{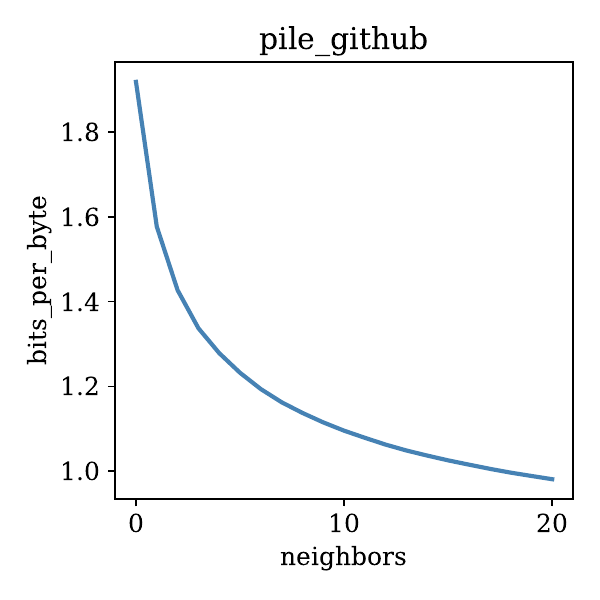}
        \label{fig:gpt2-github-bpb}
    }

    \subfloat{
        \includegraphics[width=0.13\textwidth]{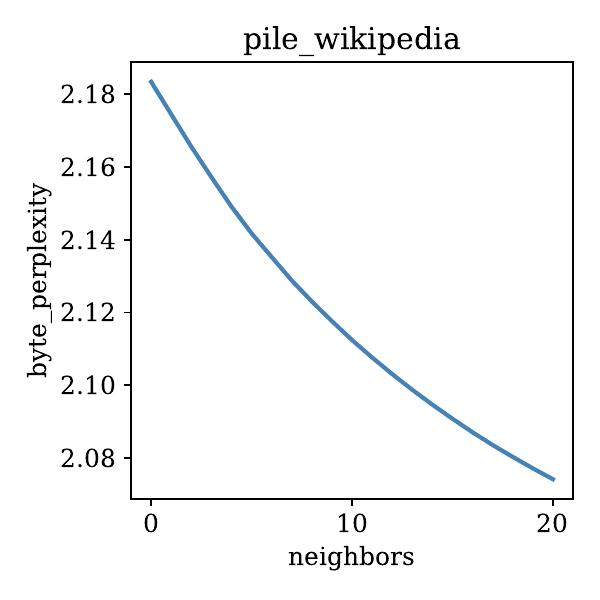}
        \label{fig:gpt-2wiki-byte_ppl}
    }
    \subfloat{
        \includegraphics[width=0.13\textwidth]{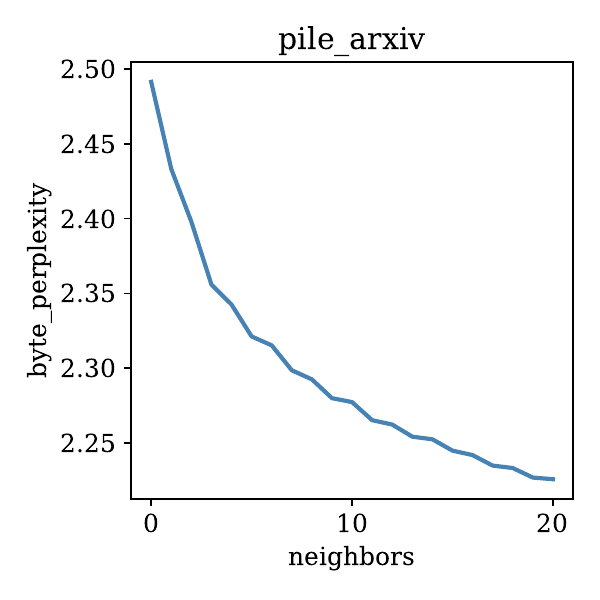}
        \label{fig:gpt2-arxiv-byte_ppl}
    }
    \subfloat{
        \includegraphics[width=0.13\textwidth]{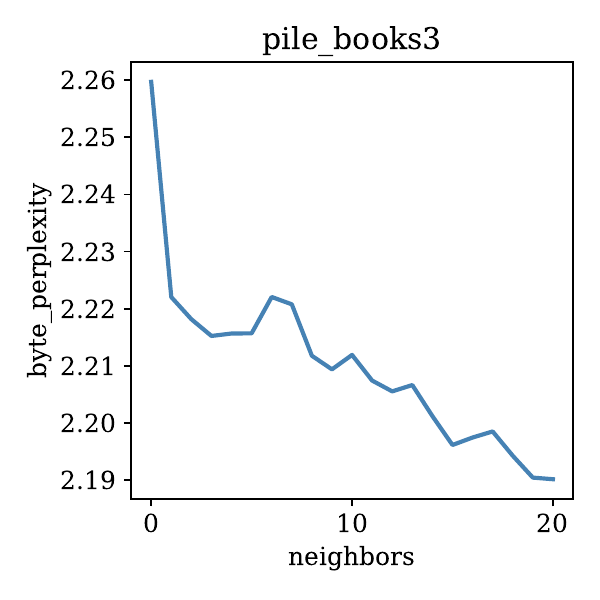}
        \label{fig:gpt2-book3-byte_ppl}
    }
    \subfloat{
        \includegraphics[width=0.13\textwidth]{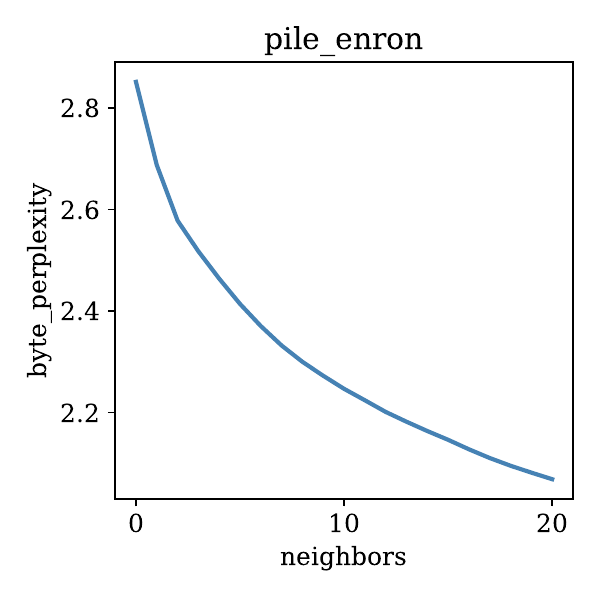}
        \label{fig:gpt2-enron-byte_ppl}
    }
    \subfloat{
        \includegraphics[width=0.13\textwidth]{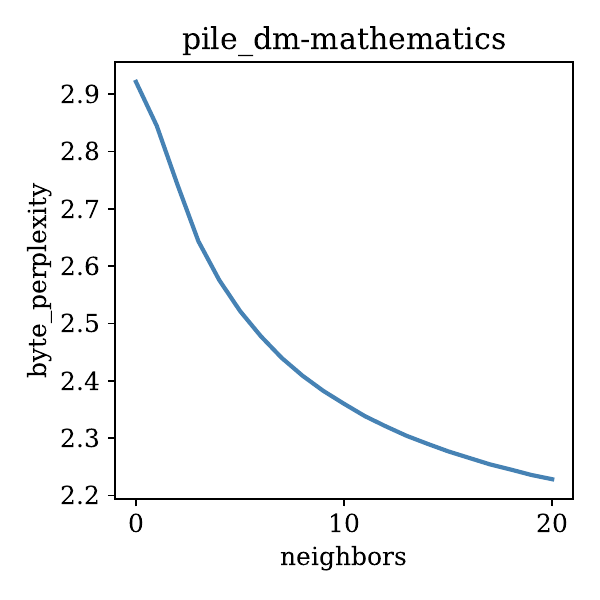}
        \label{fig:gpt2-dm-byte_ppl}
    }
    \subfloat{
        \includegraphics[width=0.13\textwidth]{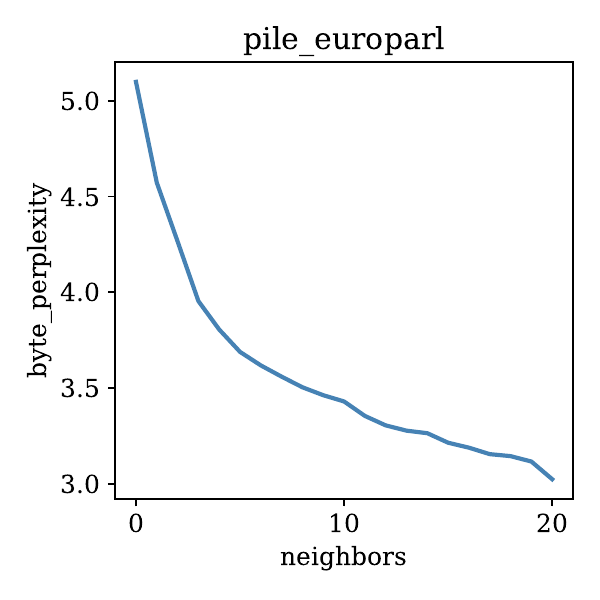}
        \label{fig:gpt2-euro-byte_ppl}
    }
    \subfloat{
        \includegraphics[width=0.13\textwidth]{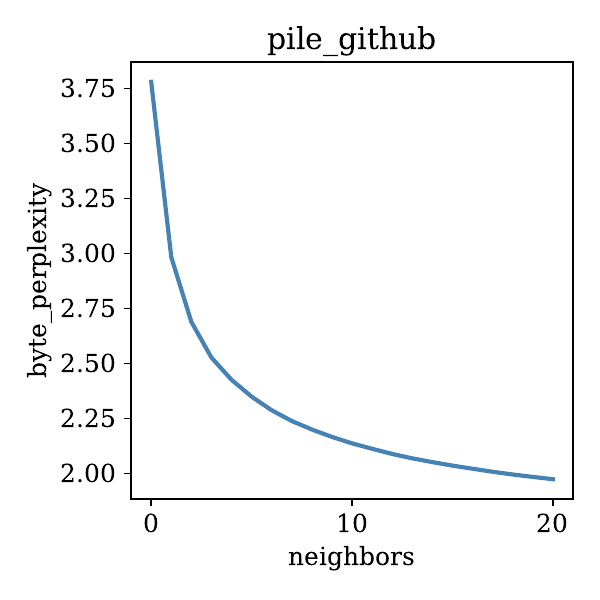}
        \label{fig:gpt2-github-byte_ppl}
    }
    \caption{Model performance steadily improves as we increase the number of neighbors used in TTT-NN on GPT-2. The first row shows perplexity, the second row shows bits per bytes, and the third row shows training loss.}
    \label{fig:gpt2_scaling_nn}
\end{figure}

\subsection{Result 1}\label{sec:res1}

We pivot on GPT-2 (117M) for the experiments to reproduce the first claim: ``metrics such as bits per byte and perplexity significantly decrease after fine-tuning on nearest neighbors''.

\paragraph{Result comparison before and after TTT-NN.} Figure~\ref{fig:gpt2-before-after} presents the bits per byte metrics before and after applying the TTT-NN approach across seven sub-datasets from The Pile using the GPT-2 model. The results consistently demonstrate significant performance improvements, validating the first claim from the original paper. The percentage improvements vary substantially across different domains. Notably, GitHub shows the most dramatic improvement with a reduction to 51\% of the before performance in bits per byte, indicating that TTT-NN is particularly effective for code-related content. This aligns with the original paper's finding that code tasks benefit substantially from this approach. Other datasets with structured or specialized content also show substantial gains: Europarl (68\%), DM-Mathematics (75\%), and Enron emails (69\%). In contrast, datasets containing more general text show more modest improvements: Wikipedia (93\%), ArXiv (88\%), and Books3 (96\%) retain a higher percentage of their original bits per byte measure. This pattern suggests that TTT-NN's effectiveness may correlate with the specificity and structure of the domain—more specialized content appears to benefit more from nearest-neighbor adaptation.

\paragraph{Scaling effects w.r.t. the number of neighbors.} Figure~\ref{fig:gpt2_scaling_nn} illustrates how model performance evolves as we progressively increase the number of nearest neighbors used for TTT-NN on GPT-2. The figure presents three key metrics: word perplexity (top row), bits per byte (middle row), and byte perplexity (bottom row). Across all datasets and metrics, we observe a consistent pattern of decreasing (performance improving) as more neighbors are added to the training process. The most dramatic improvements occur with the first few neighbors, with particularly steep performance gains visible in the specialized datasets like GitHub and Europarl. For instance, in GitHub (rightmost column), both perplexity and bits per byte show dramatic drops within the first 5 neighbors, after which the improvement curve flattens considerably. The shape of the performance curves varies by dataset type. General knowledge datasets like Wikipedia and ArXiv show smoother, more gradual improvement as neighbors increase. In contrast, the Books3 dataset exhibits a more irregular pattern with small fluctuations, suggesting that the relevance of retrieved neighbors may vary more significantly in this domain. Notably, even at 20 neighbors, most curves continue to show slight downward trends, indicating that additional neighbors beyond our experimental limit might yield further improvements. This observation aligns with the original paper's findings that while significant gains can be achieved with just 20 neighbors, scaling to 50 neighbors provides additional benefits. We observe similar patterns in reduction of training losses as we scale the number of nearest neighbors (Figure~\ref{fig:gpt2-training}). 
These results confirm that TTT-NN's effectiveness scales with the number of neighbors, though with varying efficiency across different data domains. The consistent improvement pattern across all three evaluation metrics further strengthens the validity of the approach, demonstrating that the benefits are robust across different ways of measuring language model performance.

\paragraph{Comparison to the original paper.} These results successfully reproduce the core claim from the original paper that fine-tuning on retrieved nearest neighbors significantly reduces bits per byte across diverse tasks. Our implementation using 20 nearest neighbors achieves substantial improvements, although the magnitude varies by dataset type. The consistent pattern of improvements across all tested datasets provides strong evidence for the effectiveness of the TTT-NN approach. When comparing our results with Figure 5 from the original paper, we observe similar patterns of improvement, though with some quantitative differences. The original study used 50 nearest neighbors for test-time training, while our resource constraints limited us to 20 neighbors. This difference in neighbor count explains the performance gap between our results and those reported in the original paper. For example, on the GitHub dataset, we observed a reduction to 51\% of the original bits per byte, while the original paper reported a more substantial improvement to 36\%. Similarly, on Europarl, we achieved 68\% compared to their 59\%. Despite these numerical differences, the relative pattern of improvements across datasets is consistent with the original findings. Both our study and the original paper demonstrate that specialized domains like GitHub, Europarl, and Enron show the largest relative improvements, while more general text datasets like Wikipedia and Books3 show more modest gains. This consistency in the pattern of improvements across dataset types, despite using fewer neighbors, provides strong evidence for the robustness of the TTT-NN approach and validates the original paper's conclusions about its effectiveness.

\subsection{Result 2}\label{sec:res2}

We validate the second claim: “models not pre-trained on specific
tasks can achieve comparable performance to pre-trained models after test-time training” with experiments across different GPT models, where GPT2 and GPT-2 Large were not pre-trained on The Pile dataset, while GPT-Neo was. 
\begin{figure}[tbh]
    \centering
    \subfloat[\tiny GPT2 (117M)]{
        \includegraphics[width=0.13\textwidth]{ResultsGPT2/before-after-pile_europarl.pdf}
        \label{fig:gpt2-dm}
    }
    \subfloat[\tiny GPT2-Large (774M)]{
        \includegraphics[width=0.13\textwidth]{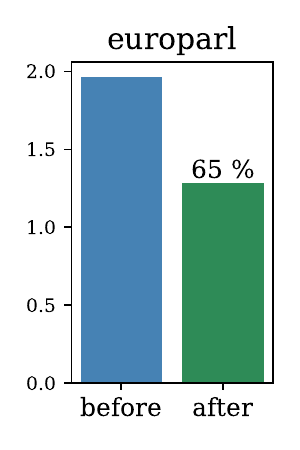}
        \label{fig:gpt2large-dm}
    }
    \subfloat[\tiny GPT-Neo (1.3B)]{
        \includegraphics[width=0.13\textwidth]{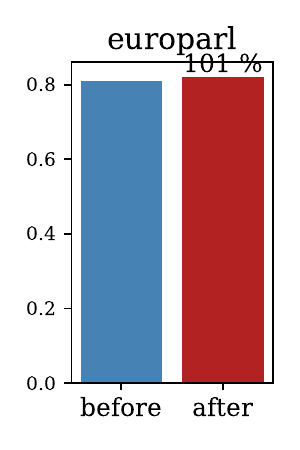}
        \label{fig:gptNeo-dm}
    }
    \subfloat[\tiny GPT2 (117M)]{
        \includegraphics[width=0.13\textwidth]{ResultsGPT2/before-after-pile_github.pdf}
        \label{fig:gpt2-dm}
    }
    \subfloat[\tiny GPT2-Large (774M)]{
        \includegraphics[width=0.13\textwidth]{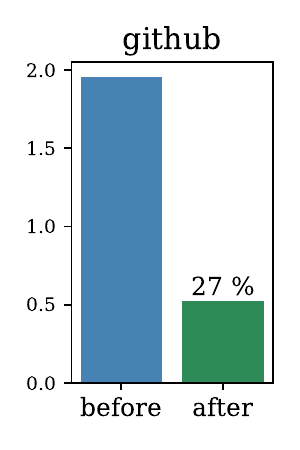}
        \label{fig:gpt2large-dm}
    }
    \subfloat[\tiny GPT-Neo (1.3B)]{
        \includegraphics[width=0.13\textwidth]{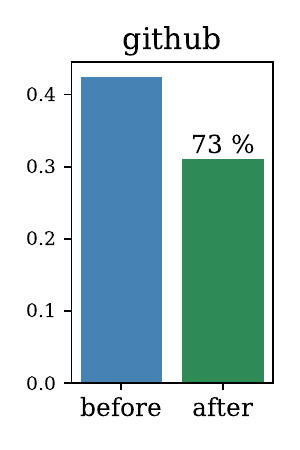}
        \label{fig:gptNeo-dm}
    }
    
    \caption{Performance comparison (bits per byte) before and after applying TTT-NN on the \textbf{EuroParl} (left) and \textbf{Github} (right) datasets using 20 nearest neighbors across models of increasing size.}
    \label{fig:model-comparison}
\end{figure}


Figure~\ref{fig:model-comparison} presents a clear pattern across both EuroParl and GitHub datasets. GPT2 (117M) and GPT2-Large (774M), which were not pre-trained on The Pile, show substantial performance improvements after applying TTT-NN, with reductions to 68\% and 65\% of original bits per byte on EuroParl, and even more dramatic improvements to 51\% and 27\% on GitHub. In contrast, GPT-Neo (1.3B), which was pre-trained on The Pile, shows performance degradation on EuroParl (113\% of original bits per byte) and more modest improvements on GitHub (73\%). Figure~\ref{fig:model-comparison-dm}, reporting results on DM-Mathematics, reveals that GPT2 (117M) and GPT2-Large (774M) benefit significantly from TTT-NN on mathematical content, reducing to 75\% and 72\% of original bits per byte respectively. In contrast, GPT-Neo (1.3B), pre-trained on The Pile, shows performance degradation (101\%). This pattern is consistent with the hypothesis in the original paper that models not previously exposed to the target distribution benefit more significantly from test-time adaptation. 


It is worth noting that the performance of GPT2-Large (774M) after TTT-NN on GitHub achieved a dramatic 73\% reduction (from 1.954 to 0.527 bits per byte, resulting in 27\% of the original value). This dramatic improvement allows GPT2-Large to achieve 0.527 bits per byte, which approaches GPT-Neo's original performance of 0.424—despite GPT-Neo (1.3B) being nearly twice the size and pre-trained on The Pile dataset. This suggests that a medium-sized model with appropriate test-time adaptation can potentially approach the performance of much larger models that were pre-trained on similar data distributions.

These results demonstrate that TTT-NN effectively bridges the gap between models pre-trained on different data distributions, allowing smaller models without specific pre-training to achieve competitive performance through efficient test-time adaptation, which confirms the original paper's claim. While our exact measurements differ slightly from those reported in the original paper (which showed GPT2-Large achieving 59\% on EuroParl, 26\% on GitHub, and 70\% on DM-Math; GPT-Neo achieving 101\%, 75\%, and 101\% respectively), the overall pattern remains consistent. We attribute these numerical differences primarily to our use of 20 nearest neighbors versus the original paper's 50, though the relative performance relationships between models are preserved across both studies.

\subsection{Additional Results not Present in the Original Paper}\label{sec:add_res}

\begin{figure}[h]
    \centering
    \scriptsize 
    \subfloat[\scriptsize GPT2 (117M)]{
        \includegraphics[width=0.14\textwidth,trim={0 0 0 0.9cm},clip]{ResultsGPT2/before-after-pile_dm-mathematics.pdf}
        \label{fig:gpt2-dm}
    }
    \subfloat[\scriptsize GPT2-Large (774M)]{
        \includegraphics[width=0.14\textwidth,trim={0 0 0 0.9cm},clip]{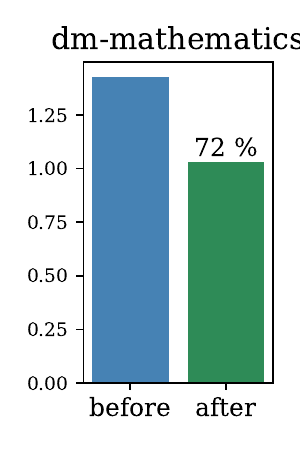}
        \label{fig:gpt2large-dm}
    }
    \subfloat[\scriptsize GPT-Neo (1.3B)]{
        \includegraphics[width=0.14\textwidth,trim={0 0 0 0.9cm},clip]{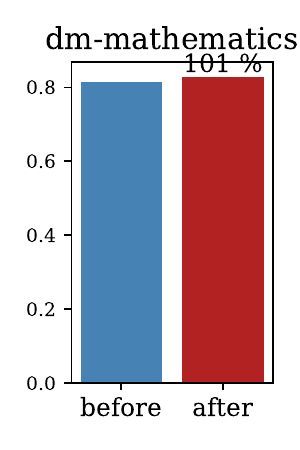}
        \label{fig:gptNeo-dm}
    }
    \subfloat[\scriptsize R1-Distilled-Qwen-1.5B]{
        \includegraphics[width=0.14\textwidth,trim={0 0 0 0.9cm},clip]{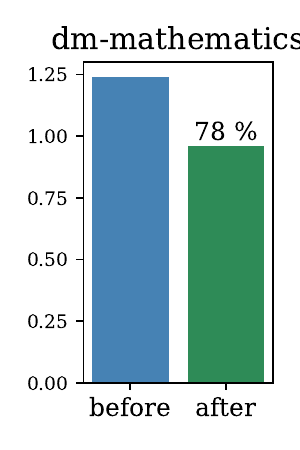}
        \label{fig:r1-dm}
    }
    \caption{ Performance comparison (bits per byte) before and after applying TTT-NN on the \textbf{DeepMind Mathematics} dataset using 20 nearest neighbors across models of increasing size.}
    \label{fig:model-comparison-dm}
\end{figure}


We evaluate TTT-NN on the R1-Distilled-Qwen-1.5B model to explore generalization to modern reasoning-focused models. As shown in Figure~\ref{fig:r1-dm}, applying TTT-NN to Qwen-1.5B on the DeepMind Mathematics dataset reduces bits per byte to 78\% of the original value, demonstrating a 22\% performance improvement.
This result contrasts notably with GPT-Neo (1.3B), which shows performance degradation at 101\% of original bits per byte, despite similar parameter counts. Meanwhile, GPT2 (117M) and GPT2-Large (774M) achieve improvements to 75\% and 72\% respectively.
The positive results for Qwen-1.5B are significant because it demonstrates that TTT-NN generalizes effectively to newer models with even stronger capabilities. Our conjecture for the improvements is that R1-Distilled-Qwen-1.5B was optimized for reasoning and chat tasks, creating a different data distribution than that of the DeepMind Mathematics dataset.
These findings suggest that model architecture and training data distribution significantly influence responsiveness to test-time adaptation.

\section{Discussion}

\subsection{What was Easy}
Reproducing several aspects of this study was straightforward due to the authors' thorough preparation of resources. The pretrained RoBERTa embeddings finetuned on The Pile dataset were readily accessible for download, eliminating the need to train these models from scratch. Starting the indexing server with a single split of the Pile dataset required minimal configuration when we can get the same computational resources, thanks to clear documentation in the codebase. Despite The Pile dataset having been removed from Hugging Face and its official website, we located an accessible copy on Hyak, which allowed us to proceed with our experiments. Additionally, the authors provided comprehensive evaluation scripts that made replicating the evaluation pipeline seamless, with metrics that matched those described in the paper.

\subsection{What was Difficult}
The most significant challenges we encountered stemmed from the substantial computational requirements described in the original paper. The authors implemented a distributed system of 30 CPU servers (1 server per split), each configured with 12 CPUs and 256 GB of RAM, totaling 360 CPUs and approximately 7.68 TB of RAM. The system is designed for instantaNeous nearest neighbor retrieval from the entire Pile dataset---a scale of infrastructure unavailable to most academic research groups without substantial cloud computing budgets. Furthermore, the codebase was designed to load entire JSONL files into memory when starting the server, requiring over 128GB RAM per instance, which is impractical with our available Hyak allocations.  Consequently, our retrieval for nearest nearest neighbor examples for each test input took 3 times longer than what was reported in the paper. However, the test-time training pipeline speed is mostly bottle-necked by one gradient update step for each sample, especially for long sequence inputs.

\subsection{Recommendations for Reproducibility}
Based on our experience, we propose several recommendations to enhance reproducibility for future work in this area:

\begin{enumerate}
    
    \item Authors may consider alternative implementations that don't require loading entire datasets into memory,  for example, our implementation of loading only the specific lines from the file into memory.
    
    \item Authors Provide Docker containers with the complete environment configuration to eliminate setup issues and ensure consistent results across different systems.
    
    \item Authors should clearly document the \textbf{minimum} and recommended computational resources needed for each experiment, including RAM, CPU/GPU specifications, storage, and \textbf{expected total run times}.
    
    \item Make the intermediate artifacts available if possible, allowing others to skip resource-intensive steps when reproducing specific parts of the pipeline. For example, saving the nearest neighbor examples per test samples.
    
    
\end{enumerate}

These recommendations would significantly lower the barrier to reproducing the impressive results reported in the paper while maintaining the integrity of the original research contribution.

\section*{Communication with Original Authors}


We did not contact the authors during our reproduction study. The paper and accompanying code repositories were sufficiently clear and well-documented to enable our reproduction efforts without requiring direct communication. The methodologies, algorithms, and implementation details were thoroughly described in the publication, and the code structure was organized in a logical manner with adequate comments. While we faced computational resource constraints in replicating the full-scale experiments, these limitations were specific to our available infrastructure and not related to any lack of clarity in the authors' materials.



\bibliographystyle{plainnat} 
\bibliography{refs}      
\newpage

\appendix
\section{Additional Results}
\begin{figure}[tbh]
    \centering
    \subfloat{
        \includegraphics[width=0.13\textwidth]{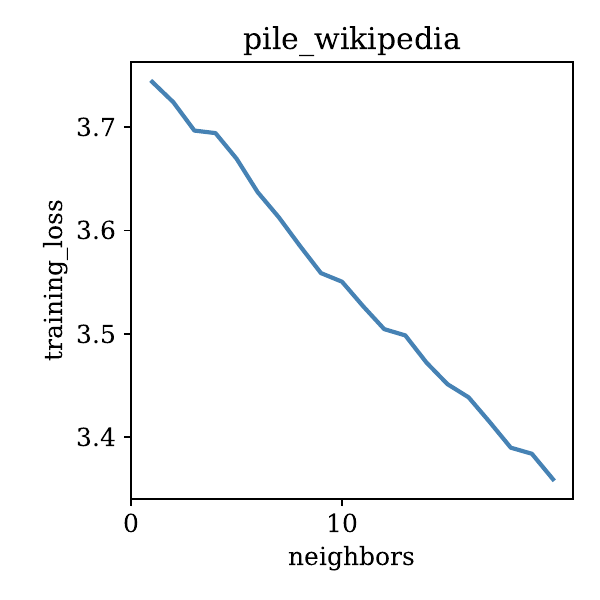}
        \label{fig:gpt-2wiki-loss}
    }
    \subfloat{
        \includegraphics[width=0.13\textwidth]{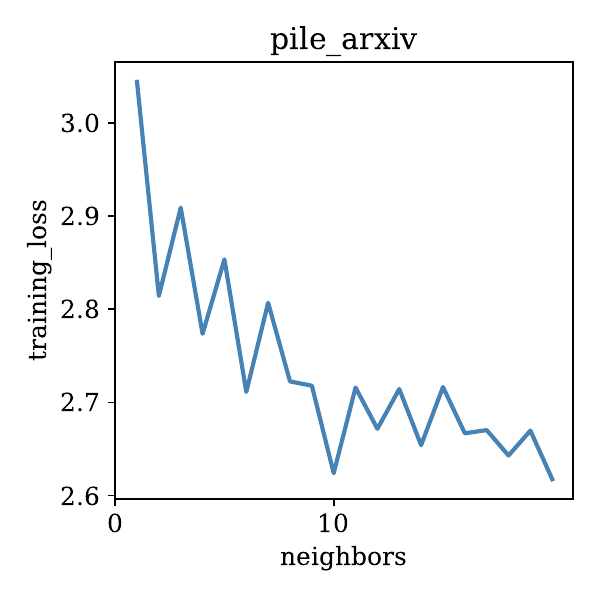}
        \label{fig:gpt2-arxiv-loss}
    }
    \subfloat{
        \includegraphics[width=0.13\textwidth]{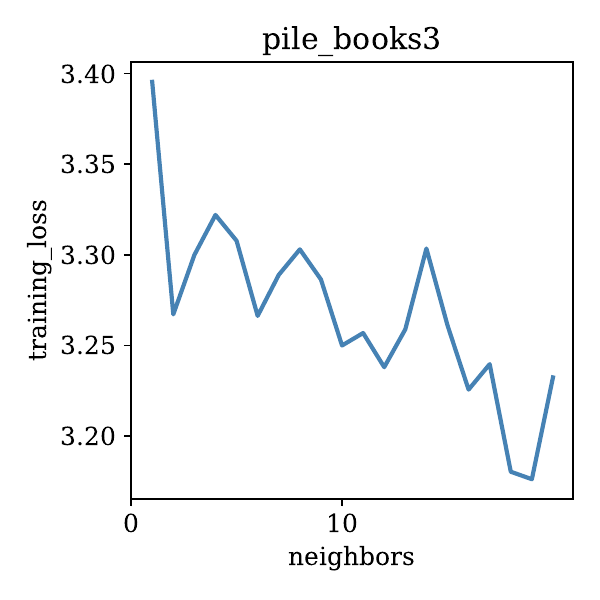}
        \label{fig:gpt2-book3-loss}
    }
    \subfloat{
        \includegraphics[width=0.13\textwidth]{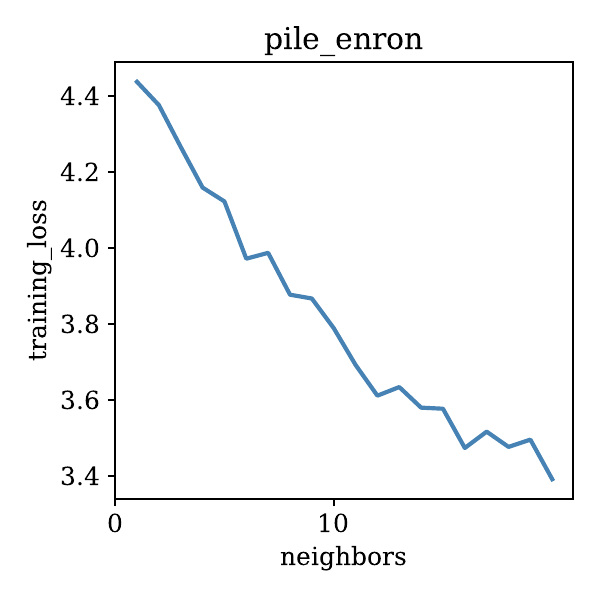}
        \label{fig:gpt2-enron-loss}
    }
    \subfloat{
        \includegraphics[width=0.13\textwidth]{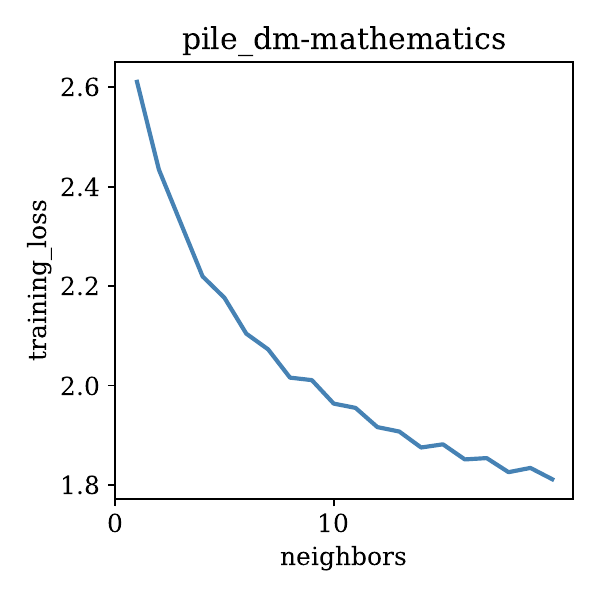}
        \label{fig:gpt2-dm-loss}
    }
    \subfloat{
        \includegraphics[width=0.13\textwidth]{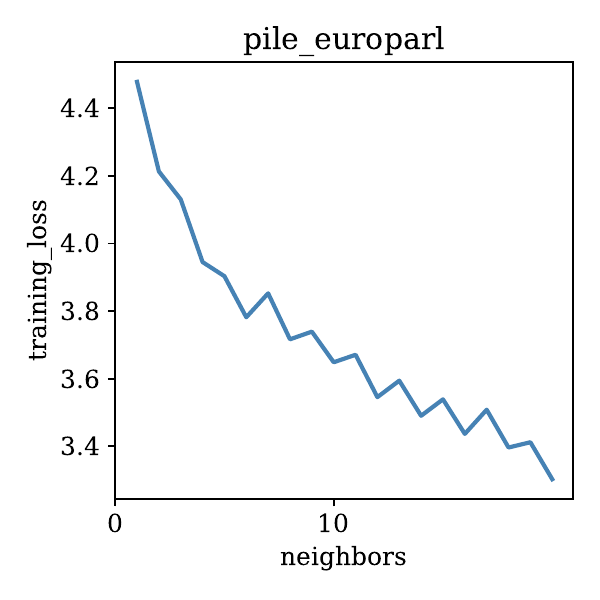}
        \label{fig:gpt2-euro-loss}
    }
    \subfloat{
        \includegraphics[width=0.13\textwidth]{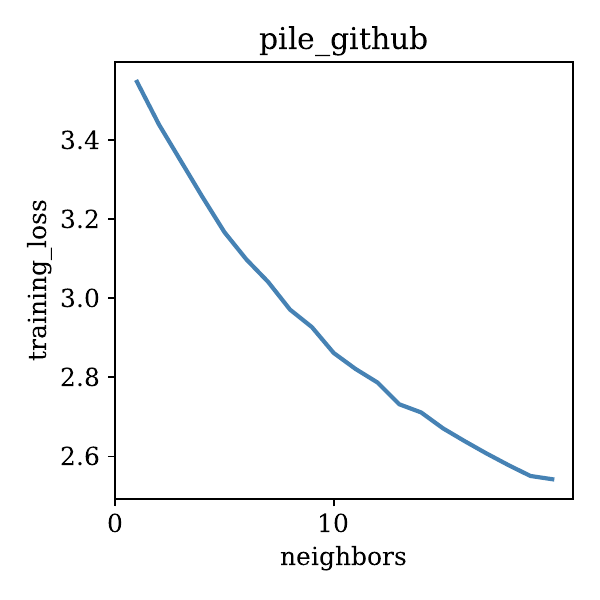}
        \label{fig:gpt2-github-loss}
    }
    
    \caption{Training losses steadily decreases as we increase the number of neighbors used in TTT-NN on GPT-2.}
    \label{fig:gpt2-training}
\end{figure}

Figure~\ref{fig:gpt2-training} demonstrates how training losses consistently decrease as more neighbors are incorporated into the test-time training process. While Wikipedia and GitHub show relatively smooth, monotonic decreases in training loss, other datasets like ArXiv, Books3, and Europarl exhibit more oscillatory behavior, with periodic fluctuations superimposed on the general downward trend. These fluctuations suggest that individual neighbors may vary in their relevance or compatibility with the test sample, occasionally introducing less optimal learning signals. Despite these variations, the overall downward trajectory of training losses across all datasets corroborates the performance improvements observed in the evaluation metrics, providing additional evidence for the effectiveness of the TTT-NN approach.

\end{document}

%% file: macro.tex
\usepackage{makecell}
\usepackage{xspace}

\usepackage{amssymb}
\usepackage{pifont}
\usepackage{etoolbox, subcaption}
\usepackage[font=footnotesize,labelfont=bf,aboveskip=0pt,belowskip=-12pt]{caption}

\newdimen\abovecrulesep
\newdimen\belowcrulesep
\makeatletter
\patchcmd{\@@@cmidrule}{\aboverulesep}{\abovecrulesep}{}{}
\patchcmd{\@xcmidrule}{\belowrulesep}{\belowcrulesep}{}{}

\definecolor{demphcolor}{RGB}{144, 144, 144}
\definecolor{mygray}{gray}{0.4}
\definecolor{lightgray}{rgb}{0.9, 0.9, 0.9}

\newlength\savewidth

\newcommand{\tablestyle}[2]{\setlength{\tabcolsep}{#1}\renewcommand{\arraystretch}{#2}\centering\footnotesize}

\renewcommand\paragraph[1]{\vspace{0.2em}\noindent\textbf{#1}}

\usepackage{etoolbox}
\preto\align{\small}
\expandafter\preto\csname align*\endcsname{\small}
\preto\equation{\par\nobreak\small\noindent}

\usepackage{enumitem}
\usepackage{multirow}